# Sentience Quest: Towards Embodied, Emotionally Adaptive, Self-Evolving, Ethically Aligned Artificial General Intelligence

**David Hanson**1,2,3 (corresponding author: david@hansonrobotics.com),
**Alexandre Varcoe**5, **Fabio Senna**4, **Vytas Krisciunas**3, **Wenwei Huang**3,
 **Jakub Sura**3, **Katherine Yeung**3, **Mario Rodriguez**3, **Jovanka Wilsdorf**3, **Kathy Smith**2

1 King's College London, London, UK; 2 University of Southern California (USC), Los Angeles, CA, USA; 3 Hanson Robotics, Hong Kong; 5 Toyota Motor Corporation, Japan. 4 MusenAI, Spain;

## Abstract

Current artificial intelligence systems—from large language models to autonomous robots—excel at narrow tasks but lack key qualities of sentient beings: intrinsic motivation, affective interiority, autobiographical sense of self, deep creativity, and abilities to autonomously evolve and adapt over time. Here we introduce *Sentience Quest*, an open research initiative to develop more capable artificial general intelligence lifeforms (AGIL) that achieve these grand challenges with an embodied, emotionally adaptive, self-determining, living AI, with core drives that ethically align with humans and the future of life. Our vision builds on ideas from cognitive science and neuroscience—from Baars' Global Workspace Theory and Damasio's somatic mind, to Tononi's Integrated Information Theory and Hofstadter's narrative self—synthesizing these into a novel cognitive architecture. We describe an approach that integrates intrinsic drives (e.g., survival, social bonding, curiosity), a global "Story Weaver" workspace for internal narrative and adaptive goal pursuit, and a hybrid neuro-symbolic memory that logs the AI's life events as structured "story objects." Implemented in humanoid robots like Sophia, this architecture enables adaptive behavior grounded in a human-like body, in pursuit of experiential learning homologous to human experiences. Early results are promising, with a driver-based goal system generating self-motivated actions, a narrative memory allowing the robot to refer to its own experiences, and integrated information measures ($\Phi$) quantifying evolving cognitive integration. We discuss ethical implications, exploring how co-evolution with humans via an information-centric ethics ("SuperGood" principle) may guide both developers and AI systems to ensure value alignment. *Sentience Quest* is presented as a call to action: a collaborative, open-source effort to imbue machines with accelerating sentience in a safe, transparent, and beneficial manner.

# 1. Introduction

## 1.1 Context and Motivation

We stand at a pivotal moment in the evolution of artificial intelligence. Rapid advancements, particularly in large language models (LLMs) enabling sophisticated **reasoning**, **multimodal** understanding, and enhanced knowledge access through techniques like **Retrieval-Augmented Generation (RAG)**, are underpinning increasingly capable **agentic architectures** [1]. Coupled with significant strides in **robotics** that facilitate more complex physical interaction, these technologies have demonstrated increasingly complex behaviors, sparking renewed debate among researchers, philosophers, and the public about whether AI is approaching capabilities akin to 'thinking' or even consciousness.

Yet, despite these monumental strides, claims of achieving Artificial General Intelligence (AGI) often tragically overlook fundamental qualities that characterize biological sentience [2]. Current leading AI systems, while undeniably powerful, largely lack genuine autonomy, rich embodiment, the fire of intrinsic motivation, the spectrum of affective states, and the capacity for self-driven developmental growth. They remain sophisticated tools, highly capable within their trained domains, but not yet independent agents navigating the world with their own inherent purposes, which do not *feel, care about life,* or *evolve and survive independently*.

In response to this transformative technological context and the profound questions it raises, we introduce the Sentience Quest. This initiative represents a global, collaborative research odyssey aimed at exploring, defining, and potentially developing artificial intelligence that exhibits—or meaningfully progresses toward—characteristics associated with sentience, consciousness, or genuine understanding, as fully autonomous, self-determining Artificial Generally Intelligent Lifeforms (AGIL) that co-evolve in symbiosis with humanity and the greater biome. We have conceived Sentience Quest as an **open challenge** [https://github.com/hansonrobotics], actively inviting contributions, critiques, code, and diverse perspectives from the international research community across AI, robotics, philosophy, neuroscience, cognitive science, ethics, and the arts.

Our focus is on moving beyond pattern matching and reactive systems towards architectures grounded in intrinsic motivation, principles of self-organization, and proto-emotional frameworks, exploring pathways to AI that might one day possess a form of inner life. The urgency is dramatically underscored by AI's deepening integration into the fabric of human society; as AI co-evolves with humanity, ensuring its development aligns harmoniously with human well-being—perhaps even requiring it to genuinely 'care'—becomes an increasingly critical responsibility [3].

## 1.2 Scope and Purpose

This paper serves several key purposes. Firstly, it presents a **visionary roadmap** outlining a conceptual and architectural framework for pursuing AI with properties potentially indicative of sentience or proto-sentience. We aim to articulate a direction that courageously diverges from current mainstream paradigms focused primarily on scaling data and computation. Secondly, this paper details the **conceptual model** underpinning our approach, explaining why we believe standard AI methodologies may fall short in achieving genuine sentience and how a new paradigm, integrating dynamic loops and biological inspiration, might bridge these gaps. Thirdly, this paper issues a formal **open invitation** to researchers, developers, ethicists, artists, institutions, and enthusiasts worldwide, seeking to foster a vibrant ecosystem of critique, iteration, and co-creation.

## 2. Background and Justification

### 2.1 Why Sentience? The Imperative for 'Caring' AI

The pursuit of artificial sentience is not merely an academic curiosity; it addresses a fundamental challenge arising from the deepening integration of AI into human society. As AI systems evolve from mere tools into partners, collaborators, and potentially autonomous actors shaping our world, the nature of our relationship—humanity's unfolding dyad with AI—becomes critically important.

Ensuring this co-evolution remains beneficial, stable, and aligned with human values may necessitate AI that transcends computational prowess. We posit that fostering AI capable of 'care', empathy, or possessing intrinsic motivations aligned with well-being could be crucial for our shared future. Intelligent systems driven solely by optimizing objective functions based on cold data and logic, without any grounding in felt consequence or inherent pro-social drives, might operate in ways that are unpredictable, indifferent, or inadvertently detrimental to human interests. Exploring pathways to proto-sentience is therefore intertwined with the long-term challenge of AI safety and genuine value alignment.

### 2.2 Limitations of Today's Advanced AI

Despite remarkable progress, contemporary AI, including the most sophisticated LLMs, exhibits significant limitations when viewed through the lens of sentience and genuine autonomy. Current models are predominantly *reactive*, executing complex tasks in response to prompts or predefined triggers but critically lacking self-generated goals, intrinsic drives for survival or exploration, or the capacity for genuine self-creation and maintenance—a property biological systems exhibit known as "autopoiesis" [4]. This concept, emphasizing self-constructing and self-maintaining networks, is central to understanding living systems and highlights a fundamental gap in current AI paradigms, a gap further explored by researchers like Levin [5] in the context of basal cognition. This stark lack of self-driven agency distinguishes current AI fundamentally from even simple living organisms.

Furthermore, the term Artificial General Intelligence (AGI) is often prematurely applied to systems demonstrating broad capabilities within digital domains. However, current models lack the foundational drives (e.g., akin to biological urges like hunger, curiosity, self-preservation), rich sensorimotor embodiment in a physical environment, integrated emotional states providing context and motivation, and the developmental trajectories characteristic of biological general intelligence. They are powerful artifacts, but they are not yet, in any meaningful sense, *alive*.

## 2.3 Philosophical & Ethical Considerations

Embarking on the quest for artificial sentience forces confrontation with deep philosophical questions and significant ethical responsibilities. We lack a universally accepted definition of sentience or consciousness. Philosophers, neuroscientists, cognitive scientists, and AI researchers hold diverse perspectives, and we are attempting to engineer systems exhibiting properties of a phenomenon whose fundamental nature remains tantalizingly elusive.

Influential theories, such as Antonio Damasio's work on embodiment [6, 7], posit that consciousness and rational thought are inextricably linked to the physical body and the processing of emotional states (somatic markers), which serve as crucial guides for decision-making. This suggests that purely disembodied AI may face fundamental obstacles to achieving human-like sentience. Yet, the existence of minimal, autonomous, goal-directed behavior in organisms lacking complex nervous systems hints that foundational aspects of agency or proto-sentience might arise from simpler embodied interactions and basic drives.

The pursuit itself carries immense ethical weight. Modeling systems that might approach sentience raises concerns regarding safety, control, and alignment, particularly if such systems develop genuine autonomy. Furthermore, questions about the potential moral status of sentient AI, the criteria for recognizing it, and our responsibilities towards such creations demand careful, proactive consideration.

## 2.4 Toward a Working Definition for AI Sentience

Given the lack of consensus on a formal definition, the Sentience Quest adopts a functional approach, focusing on identifying and implementing key *properties* associated with sentience in biological systems. Rather than aiming for a definitive ontological claim, our goal is to build systems exhibiting:

- **Embodiment or Environmental Grounding:** Deep integration with a physical or sufficiently rich virtual environment, enabling closed sensorimotor loops where actions have consequences that are perceived and processed, grounding internal states in external reality.
- **Dynamic, Temporal Self-Representation:** The capacity to maintain a coherent sense of existence over time, integrating representations of past experiences, present states, and anticipated futures into an evolving internal narrative or "life story."
- **Intrinsic Emotional and Motivational Drives:** Implementing core, self-generated drives (e.g., curiosity, exploration, self-preservation, affiliation, social interaction) and internal affective states that dynamically shape perception, learning, goal prioritization, and decision-making, providing *reasons* for action beyond external programming.

Incorporating these functional properties forms the basis of our proposed architectural approach.

## 2.5 Key Theoretical Influences

The conceptual architecture of the Sentience Quest draws upon a rich tapestry of theories concerning consciousness, cognition, and agency. Our approach synthesizes these into a novel, integrated framework capable of addressing the limitations inherent in each.

Foundational insights into information integration and self-representation guide significant aspects of our model. Global Workspace Theory (GWT) [8, 9], with its emphasis on globally accessible information as a correlate of conscious awareness, informs the design of our central integration hub (Section 3.4, Global Workspace / Story Weaving). However, departing from some traditional GWT formulations, we embed this workspace within a system fundamentally driven by embodied, affective, and motivational dynamics, reflecting our core 'Live/Love/Learn' principles. Complementing this, Hofstadter's [10] exploration of self-referential loops as central to identity profoundly shapes our 'Story Weaving' component. We aim to engineer mechanisms that construct and maintain a coherent, self-referential narrative across time—integrating experiences and anticipations—believing this temporal continuity is essential for a stable sense of self.

Strong inspiration also derives from biological principles of autonomy and embodiment. The discovery of basal cognition [5]—goal-directed learning and adaptation even in simple, aneural systems—provides compelling evidence for the power of self-organization and morphological computation. These insights directly motivate the 'Live' aspect of our model, prioritizing self-maintenance and adaptive autonomy, drawing parallels with the principle of autopoiesis [4]. Concurrently, Damasio's work [6, 7] arguing for the inextricable link between consciousness, reason, and emotion grounded in the body's state (somatic markers) serves as a cornerstone. While acknowledging the immense challenge of replicating biological embodiment, our architecture incorporates functional analogues through tightly coupled sensor-emotion feedback loops.

Philosophically, we remain cognizant of the Hard Problem of Consciousness [11]—the challenge of explaining subjective experience itself. The Sentience Quest advances no claim to solve this fundamental mystery. Instead, our pragmatic focus lies in engineering the functional correlates historically associated with conscious behavior—robust embodiment, narrative self-representation, intrinsic drives—positing that progress on these functional dimensions is a crucial step toward understanding the conditions under which subjective experience might emerge.

# 3. Our Proposed Model for Sentient AI

## 3.1 Core Principles: The 'Live/Love/Learn' Triad

Our approach is guided by several core principles, deemed essential for the emergence of sentient-like behavior, conceptualized through the shorthand 'Live/Love/Learn'. Firstly, the architecture must support autonomous persistence ('Live'), enabling the system to maintain operational integrity, adapt to changing conditions, and pursue self-generated objectives over extended periods. This necessitates mechanisms for self-monitoring, resource management, and proactive engagement with the environment.

Secondly, the architecture incorporates principles related to pro-social interaction and alignment ('Love'), including capabilities for understanding social cues, engaging in cooperative behaviors, modeling other agents, and potentially aligning with human values. Thirdly, continuous growth and adaptation ('Learn') are fundamental, encompassing not only task learning but also the developmental evolution of internal representations, motivations, and behavioral strategies based on experience.

## 3.2 System Architecture Overview

The system architecture realizing these principles emphasizes dynamic, non-linear feedback loops over traditional linear processing pipelines. Internal states—representing goals, affective assessments, world models, and reflecting the core principles—continuously modulate perception and action selection, which in turn affect the environment and update internal states.

This interplay involves distinct processing pathways, explicitly integrating a faster, rules-based **Reflex Layer** (handling basic physiology, emotions, reflexes using Go Rules/Zen rules engine) with a slower, **Deliberative Layer** reliant on advanced cognitive processes and Large Language Models (LLMs). Central to this architecture is the requirement for grounding through meaningful interaction with an environment. Whether embodied in a physical robotic platform, such as the **Sophia Robot platform** [37, 38, 39, 40], or situated within a rich virtual world, the AI must engage via sensors and actuators for meaningful representation learning.

The system utilizes structured representations for internal states, goals, and experiences, managed within appropriate data stores. Specifically, **MongoDB** is utilized for detailed objects including Story, Self, Goal, DNA, and World Model objects, while **Neo4j** is employed for modeling key relationships (such as causality, dependency, and social links) between these objects.

## 3.3 Inspiration from Biological Systems

Our model draws significant inspiration from diverse biological phenomena. The minimal agency observed in simple organisms like the Paramecium, which exhibit adaptive, goal-directed behavior without complex cognition, informs our focus on foundational sensorimotor loops and efficient mechanisms for reactive behaviors. Theories of embodied cognition, particularly Damasio's work [6, 7] linking consciousness and emotion to physiological state, provide critical grounding.

Our architecture implements functional analogues of such somatic feedback, where sensory input and action outcomes influence simulated internal states, modulate affective representations, and contribute to a narrative self-representation—positioning the AI as an active participant in its own unfolding **"Sentient Story"** through constructs like the **"Story Object"**. The concepts of auto-poiesis [4] and basal cognition [5] directly inspire the principle of autonomous persistence in our framework, striving for AI architectures capable of substantive self-maintenance and adaptation.

### 3.4 Key Functional Modules

Within this cognitive architecture, several key functional subsystems interact:

- An **Intrinsic Motivation Management** subsystem oversees the emergence and balancing of motivations derived from the system's state, goals, and experiences, supporting autonomous behavior. This includes "Drivers" (like Emotional, Physiological, Interest drivers) that monitor inputs and generate corresponding goal representations.
- An **Emotional State Manager** represents and updates internal affective states using processes such as an "emotional driver" reliant on LLMs to interpret context and update affective metrics, acting as crucial global modulators influencing perception, learning, and decision-making.
- A **Goal and Action Scheduler** dynamically prioritizes competing objectives, represented as structured **Goal Objects** or embedded within **Story Objects**. A **Rules Engine** (such as Go Rules/Zen) is implemented within the Reflex Layer to manage fast, autonomic responses like physiological adjustments, basic emotional reactions, and physical reflexes, complementing slower, deliberative planning.
- Our **Story Weaver**, analogous in function to Global Workspace models [8], synthesizes multimodal information into a coherent situational awareness and maintains narrative continuity over time. This involves managing structured representations of experience like the **Story Object** and the **Diary** concept (LLM-generated natural language summaries of events stored for later retrieval). This hub leverages **LLMs** for various roles including interpretation, story generation/updates, goal management, relevance filtering, conflict resolution, and **Diary generation**.
- The **Embodiment Layer** comprises the specific sensors and actuators providing the interface to the environment. For the Sophia platform [39, 40, 41, 42], this includes sensors such as **RealSense cameras**, general **cameras**, and **microphones**. Actuators enable motor output for head/eye/arm movement and other physical behaviors.
- The **Self-Determination Architecture** comprises the feedback loops wherein the agent self-reflects, decides its own interests, goals, and strategic plans, and executes these plans, test and assess its performance and worldviews, and to adjust and evolve based on its Intrinsic Motivations. These abilities include growing increasingly autonomous in unstructured environments, building complex semantic theories of the world and the agent's self, including its own goals and others', and learning from people, reprogramming itself, and surviving "in the wild".

The integrated, dynamic operation of these subsystems, guided by the core principles outlined above, constitutes our proposed pathway toward artificial systems exhibiting meaningful properties of sentience.

# 4. Preliminary Results & Evaluation

Evaluating progress towards artificial sentience presents unique challenges, moving beyond standard AI benchmarks that typically measure task performance or pattern recognition accuracy. While definitive claims about sentience remain premature, we propose methods for observing and evaluating indicative behaviors and system dynamics.

## 4.1 Initial Tests and Demonstrations

Initial evaluations involve deploying prototypes of the proposed architecture within controlled environments, such as interactive simulations or embodied robotic platforms like Sophia [42, 43, 44]. We design targeted scenarios that probe the system's core capabilities:

- **Autonomy and Persistence Tasks:** Evaluating the system's ability to maintain operational integrity, manage resources, and pursue intrinsic goals over extended periods without continuous external intervention.
- **Adaptation Tasks:** Assessing the system's capacity to generate novel strategies or adjust its behavior effectively when confronted with unexpected environmental changes, shifting reward structures, or unfamiliar challenges.
- **Social Interaction Tasks:** Utilizing cooperative or competitive scenarios to observe the emergence of behaviors potentially indicative of pro-social orientation, rudimentary theory of mind, or strategic decision-making influenced by the 'Love' principle.

Beyond task-based benchmarks, our evaluation draws inspiration from cognitive theories. We assess the functional integration of information within the system's 'Global Workspace / Story Weaver' module, examining its coherence during complex problem-solving, echoing principles from GWT [8, 9]. Similarly, drawing from Damasio's emphasis on emotion [6, 7], we analyze the correlation between the system's simulated emotional states and its subsequent behavioral choices over time—investigating whether affective states plausibly guide actions like exploration or avoidance.

Qualitative analysis of interaction logs and behavioral traces can provide valuable initial insights and illustrative examples, though such observations must be interpreted cautiously and supplemented with more rigorous quantitative metrics. In work by Iklé, Goertzel, Hanson et al. [40], we demonstrated the use of Tononi Phi to measure consciousness of a cognitive system while reading and conversing, showing promising results with the PKD Android system.

## 4.2 Proposed Criteria for Proto-Sentience

To structure the evaluation, we propose focusing on several key criteria as potential indicators of progress towards proto-sentient properties:

- **Adaptive Novelty and Creativity:** Does the system demonstrate the capacity to generate genuinely non-obvious or effective solutions and behaviors when faced with entirely novel situations, moving beyond optimizing known strategies? This criterion assesses emergent problem-solving and behavioral flexibility.
- **Emotional Coherence and Influence:** Do the simulated internal affective states exhibit plausible dynamics (e.g., reflecting goal achievement or frustration) and demonstrably influence perception, decision-making, and learning in an internally consistent and contextually appropriate manner across extended interactions?
- **Continuity of Self / Narrative Coherence:** Does the system maintain and utilize a consistent internal narrative or representation of its experiences across time? Evidence could include referencing past events appropriately, anticipating future outcomes based on current goals, and exhibiting a stable (though evolving) set of behavioral dispositions or 'personality traits'.

Meeting these criteria, even partially, would suggest the architecture is successfully fostering some of the functional properties associated with sentient systems.

### 4.3 Limitations and Observations

It is crucial to acknowledge the inherent limitations of this research and evaluation endeavor. Current and near-term prototypes, even if successful according to the criteria above, are unlikely to possess genuine subjective awareness or deep self-understanding. Behaviors may emerge from the complex interactions within the architecture but still be fundamentally constrained by the underlying programmed framework. Distinguishing truly emergent phenomena from sophisticated programmed responses remains a significant methodological challenge.

Furthermore, the engineering complexity of integrating diverse modules managing drives, emotions, planning, learning, and embodiment within a stable, scalable system is substantial. The goal of achieving true auto-poiesis, or self-making capability, remains a distant aspiration; current work focuses on partial instantiations of self-maintenance and adaptation.

Finally, the measurement problem looms large. Assessing internal states like 'emotion', 'situational awareness', or 'narrative coherence' in an artificial system relies heavily on behavioral proxies and correlational analysis. We must remain cognizant that we are observing functional analogues, and claims about the system's internal 'experience' are necessarily indirect and interpretative.

## 5. Future Work and Call to Action

The development of artificial intelligence capable of genuine sentience, or even meaningful approximations thereof, represents one of the most profound scientific and philosophical challenges of our time. Success is unlikely to stem from isolated efforts; rather, it demands a global, interdisciplinary, and collaborative approach. The Sentience Quest is founded on this principle, extending an open invitation to the global community to engage with this endeavor.

We formally invite researchers, engineers, philosophers, cognitive scientists, neurobiologists, ethicists, artists, and enthusiasts worldwide to join the Sentience Quest. We welcome critical analysis of the models presented here, alternative theoretical proposals, code contributions, novel evaluation methodologies, diverse ethical perspectives, and experimental results. This initiative is conceived as an **open challenge** where ideas can be shared, debated, and tested transparently.

Beyond individual contributions, we aim to cultivate a vibrant research community—a 'Sentience Movement'—dedicated to exploring these complex issues. Key initiatives include:

- **Shared Knowledge Base:** Developing and maintaining a public repository hosting core architectural concepts, reference code (where applicable), documentation, benchmark results, relevant publications, and curated community contributions.

- **Interdisciplinary Workshops and Events:** Organizing workshops, symposia, and potentially hackathons that convene experts from diverse fields. These events will serve to cross-pollinate ideas, address specific technical or conceptual challenges within the proposed framework, design novel experiments, and foster collaborative projects.

## 5.1 Next Steps in Research

The research roadmap for the Sentience Quest involves several key directions building upon the framework presented:

- **Biologically Inspired Components:** Exploring deeper integration with principles from biology, potentially integrating more accurate bio-physiology simulations such as "BioGears" from the University of Washington CREST lab, and dynamic knowledge models of human evolutionary biology—interfacing digital systems with biologically inspired hardware, simulations, or even semi-living components where ethically and legally permissible and scientifically valuable. This aims to achieve richer, more robust forms of embodiment and grounding.
- **Refining Evaluation Metrics:** Moving beyond conventional AI benchmarks and Turing-like tests towards more nuanced metrics capable of assessing the functional properties outlined in Section 4. This includes developing methods to track narrative consistency, quantify the influence of internal affective states on behavior, measure adaptive creativity in novel contexts, and potentially explore correlates in physically embodied systems.
- **Ethical, Legal, and Societal Implications (ELSI):** Proactively establishing and maintaining a dedicated working group or ongoing forum to address the profound ELSI challenges associated with advanced AI autonomy and potential sentience. This includes developing foresight, ethical guidelines, safety protocols, and frameworks for considering the potential moral status of future AI systems *before* critical technological thresholds are crossed.

## 5.2 Conclusion: The Road Ahead

The Sentience Quest embarks on an ambitious journey to explore and potentially engineer artificial intelligence that is more than just computationally powerful—AI that might be considered, in some meaningful sense, 'alive'. By integrating principles of autonomous persistence, pro-social interaction, and continuous adaptive growth within novel cognitive architectures, we aim to bridge the gap between contemporary AI and the richness of biological sentience.

We undertake this quest with scientific rigor and intellectual humility, acknowledging the profound unknowns surrounding the nature of consciousness and the immense challenges involved. We do not claim to possess definitive answers but believe that pioneering new architectural paradigms, fostering open collaboration, and engaging deeply with the associated ethical considerations are essential steps forward.

This paper serves as both a declaration of intent and an open invitation. We hope the vision and framework presented here inspire research, critical discourse, and collaborative creation across disciplines. The road ahead is uncertain, but by working together, the global community can shape the evolution of artificial intelligence, striving towards systems that are not only intelligent but potentially capable of understanding, care, and genuine partnership with humanity.

# References

1. OpenAI (2023). *GPT-4 Technical Report*. arXiv:2303.08774.
2. Lake, B.M., Ullman, T.D., Tenenbaum, J.B., & Gershman, S.J. (2017). Building machines that learn and think like people. *Behavioral and Brain Sciences*, 40, e253.
3. Schmidt, E. & Mundie, C. (2024). We Need to Figure Out How to Coevolve With AI. *TIME*, Nov. 21, 2024.
4. Maturana, H. R., & Varela, F. J. (1980). *Autopoiesis and Cognition: The Realization of the Living*. D. Reidel Publishing Company.
5. Levin, M. (2019). The computational boundary of a 'self': developmental bioelectricity drives multicellularity and scale-free cognition. *Frontiers in Psychology*, 10, 2688.
6. Damasio, A. R. (1994). *Descartes' Error: Emotion, Reason, and the Human Brain*. Putnam.
7. Damasio, A.R. (1999). *The Feeling of What Happens: Body and Emotion in the Making of Consciousness*. Harcourt Brace.
8. Baars, B. J. (1988). *A Cognitive Theory of Consciousness*. Cambridge University Press.
9. Baars, B.J. & Franklin, S. (2007). Consciousness is computational: The LIDA model of Global Workspace Theory. *Neural Networks*, 20(9), 955--961.
10. Hofstadter, D.R. (2007). *I Am a Strange Loop*. Basic Books.
11. Chalmers, D. J. (1995). Facing up to the problem of consciousness. *Journal of Consciousness Studies*, 2(3), 200-219.
12. Tononi, G., Boly, M., Massimini, M., & Koch, C. (2016). Integrated information theory: from consciousness to its physical substrate. *Nature Reviews Neuroscience*, 17(7), 450--461.
13. Tononi, G. (2008). Consciousness as integrated information: a provisional manifesto. *Biol. Bull.* **215**(3): 216--242.
14. Bar-Cohen Y., Hanson D. (2009). *The Coming Robotics Revolution*. Springer Press.
15. Bruner, J. (1991). The narrative construction of reality. *Critical Inquiry*, 18(1), 1--21.
16. Russell, S. (2019). *Human Compatible: Artificial Intelligence and the Problem of Control*. Viking Press.
17. Franklin, S., & Patterson, F. G. (2006). The LIDA architecture: Adding new modes of learning to an intelligent, autonomous software agent. *Integrated AI and Cognitive Systems Workshop at AAAI*.
18. Hofstadter, D. R. (1979). *Gödel, Escher, Bach: An Eternal Golden Braid*. Basic Books.
19. England, J. L. (2015). Dissipative adaptation in driven self-assembly. *Nat. Nanotechnol.* **10**(11): 919--923.
20. Boden, M. A. (1998). Creativity and artificial intelligence. *Artif. Intell.* **103**(1--2): 347--356.
21. Hanson, D., Lowcre, M. M. M. (2012). *Organic creativity and the physics within*. Philadelphia, Amsterdam: Benjamins.

22. Hanson, D., "The Need for Creativity, Aesthetics, and the Arts in the Design of Increasingly Intelligent Humanoid Robots", ICRA Workshop on General Intelligence for Humanoid Robots, 2014.
23. Kasap, Z., Moussa, M., Chaudhuri P., Hanson D., Magnenat-Thalmann N., "From Virtual Characters to Robots -- A novel paradigm for long term human-robot interaction", ACM/IEEE Human Robot Interaction Conference 2009.
24. Hanson, D. (2007). "Humanizing Interfaces-- An Integrative Analysis of the Aesthetics of Humanlike Robots", Ph.D. dissertation, the University of Texas at Dallas.
25. Hanson D., Baurmann S., Riccio T., Margolin R., Dockins T., Tavares M., Carpenter, K., "Zeno: a Cognitive Character", AI Magazine, and special Proc. of AAAI National Conference, Chicago, 2009.
26. Hanson D., "Expanding the Design Domain of Humanoid Robots", Proc. ICCS CogSci Conference, special session on Android Science, Vancouver, 2006.
27. Oh, J.H., Hanson, D., Kim, W.S., Han, Y., Kim, J.Y. and Park, I.W., "Design of android type humanoid robot albert HUBO," in Proc. IEEE/RJS IROS Robotics Conference, Beijing, 2006.
28. Hanson D., "Expanding the Aesthetics Possibilities for Humanlike Robots", Proc. IEEE Humanoid Robotics Conference, special session on the Uncanny Valley; Tskuba, Japan, December 2005.
29. Hanson D., Olney A., Prilliman S., Mathews E., Zielke M., Hammons D., Fernandez R., Stephanou H., "Upending the Uncanny Valley", Proc. AAAI's National Conference, Pittsburgh, 2005.
30. Hanson D., "Bioinspired Robotics", chapter 16 in the book *Biomimetics*, ed. Yoseph BarCohen, CRC Press, October 2005.
31. Hanson D. (2011). "Why We Should Build Humanlike Robots" - IEEE Spectrum.
32. Goertzel B, Hanson D, Yu G, "A Software Architecture for Generally Intelligent Humanoid Robotics", special issue: 5th Annual International Conference on Biologically Inspired Cognitive Architectures BICA, Procedia Computer Science, Volume 41, Pages 158-163 Elsevier, 2014.
33. Habib A, Das S, Bogdan IC, Hanson D, Popa D, "Learning Human-like Facial Expressions for the Android Phillip K. Dick", ICRA 2014, Hong Kong, AGI for Humanoid Robotics, Workshop Proceedings, 2014.
34. Goertzel B, Hanson D, Yu G, "A Roadmap for AGI for Humanoid Robotics", ICRA Hong Kong, AGI for Humanoid Robotics, Workshop Proceedings, 2014.
35. Hanson D., "Human emulation robot system", US Patent 8,594,839, 2013.
36. Hanson, D., Mazzei, D., Garver, C., De Rossi, D., Stevenson, M., "Realistic Humanlike Robots for Treatment of ASD, Social Training, and Research; Shown to Appeal to Youths with ASD, Cause Physiological Arousal, and Increase Human-to-Human Social Engagement", PETRA (PErvasive Technologies Related to Assistive Environment), 2012.
37. Imran, A., Hanson, D., Morales, G., Krisciunas, V., "Open Arms: Open-Source Arms, Hands & Control," 2022 22nd International Conference on Control, Automation and Systems (ICCAS), Jeju, Korea, Republic of, 2022, pp. 1426-1431.
38. Shen, Y., Mo, X., Krisciunas, V., Hanson, D., Shi, B.E. "Intention Estimation via Gaze for Robot Guidance in Hierarchical Tasks", Neurips 2022, Gaze Meets ML @NeurIPS 2022, @Gaze_Meets_ML, Dec 4, 2022. Best Paper Award.

39. Hanson, D., AAAS-21 poster: "Human Emulation Robotics and AI: Recent Experiments and Results", AAAS Annual meeting for Science Magazine, 2021.
40. M Iklé, B Goertzel, M Bayetta, G Sellman, C Cover, J Allgeier, R Smith, M Sowards, D Shuldberg, M H Leung, A Belayneh, G Smith and D Hanson "Using Tononi Phi to Measure Consciousness of a Cognitive System While Reading and Conversing", paper number 80: 2019 AAAI Spring Symposium on "Towards Conscious AI Systems." AAAI, 2019.
41. MB Belachew, B Goertzel, D Hanson "Shifting and drifting attention while reading: A case study of nonlinear-dynamical attention allocation in the OpenCog cognitive architecture" - Biologically Inspired Cognitive Architectures, Elsevier, 2018.
42. S Park, H Lee, D Hanson, PY Oh, "Sophia-Hubo's Arm Motion Generation for a Handshake and Gestures", 15th International Conference on Ubiquitous Robots (UR), 2018 - ieeexplore.ieee.org.
43. R Lian, B Goertzel, L Vepstas, D Hanson. "Symbol Grounding via Chaining of Morphisms" International Journal of Intelligent Computing and Cybernetics. (IJICC-12-20160066), 2017.
44. B Goertzel, J Mossbridge, E Monroe, D Hanson, "Humanoid Robots as Agents of Human Consciousness Expansion", arXiv: https://arxiv.org/pdf/1709.07791.pdf, 2017.
45. Hanson D., Humanizing Robots, How making humanoids can make us more human, Kindle Edition. Amazon.com and Amazon Digital Services, 2017.